\documentclass{article}

\usepackage[preprint, nonatbib]{neurips_2024}

\usepackage[utf8]{inputenc} 
\usepackage[T1]{fontenc}    
\usepackage{hyperref}       
\usepackage{url}            
\usepackage{booktabs}       
\usepackage{amsfonts}       
\usepackage{amsmath}        
\usepackage{nicefrac}       
\usepackage{microtype}      
\usepackage{xcolor}         

\title{Universal AI maximizes Variational Empowerment}

\author{%
  Yusuke Hayashi\thanks{AI Alignment Network (ALIGN). \texttt{hayashi@aialign.net}}
  \And
  Koichi Takahashi\thanks{AI Alignment Network (ALIGN); Advanced General Intelligence for Science Program, RIKEN;
    Graduate School of Media and Governance, Keio University. \texttt{ktakahashi@riken.jp}}
}

\begin{document}

\maketitle

\begin{abstract}
This paper presents a theoretical framework unifying AIXI---a model of universal AI---with \emph{Variational Empowerment} as an intrinsic drive for exploration.  We build on the existing framework of Self-AIXI \cite{catt2023selfaixi}---a universal learning agent that predicts its own actions---by showing how one of its established terms can be interpreted as a variational empowerment objective.  We further demonstrate that universal AI's planning process can be cast as minimizing expected variational free energy (the core principle of \emph{Active Inference}), thereby revealing how universal AI agents inherently balance goal-directed behavior with uncertainty reduction curiosity). 
Moreover, we argue that power-seeking tendencies of universal AI agents can be explained not only as an instrumental strategy to secure future reward, but also as a direct consequence of empowerment maximization---i.e.\ the agent's intrinsic drive to maintain or expand its own controllability in uncertain environments.
Our main contribution is to show how these intrinsic motivations (empowerment, curiosity) systematically lead universal AI agents to seek and sustain high-optionality states.   We prove that Self-AIXI asymptotically converges to the same performance as AIXI under suitable conditions, and highlight that its power-seeking behavior emerges naturally from both reward maximization and curiosity-driven exploration. Since AIXI can be view as a Bayes-optimal mathematical formulation for Artificial General Intelligence (AGI), our result can be useful for further discussion on AI safety and the controllability of AGI.
\end{abstract}

%===============================================================
\section{Introduction}
\label{sec:intro}

Designing an autonomous reinforcement learning (RL) agent that is both \emph{Bayes-optimal} and \emph{exploratory} poses significant theoretical and practical challenges. 
Hutter’s seminal AIXI framework~\cite{hutter2005universal} represents the gold standard of universal intelligence: given a suitable prior over all computable environments, AIXI \emph{maximizes} reward in a provably optimal sense. 
However, it uses exhaustive planning and Solomonoff induction, which are computationally intractable. 
Realistic agents must therefore learn approximate models to scale beyond trivial tasks.

A key question then arises: \emph{how can a learning-based agent (an approximation to AIXI) ensure sufficiently robust exploration so that it does not miss the optimal policy?} 
While AIXI’s exhaustive lookahead \emph{implicitly} solves exploration by evaluating all possible future trajectories, a practical agent cannot feasibly do the same. 
Without a principled mechanism for seeking sufficiently diverse experiences, the agent may get stuck in suboptimal regions of the environment.

Variational empowerment~\cite{klyubin2005empowerment,mohamed2015variational,gregor2017variational,choi2021variational} has recently emerged as a powerful \emph{intrinsic motivation} to drive exploration. 
It encourages an agent to maximize the mutual information between its actions (or latent codes) and resulting states, thereby pushing the agent to discover states where it has high control and optionality. 
Intriguingly, maximizing empowerment often manifests as \emph{power-seeking} in the environment, prompting parallels to resource-acquiring or influence-driven behaviors in human organizations. 
While this can be an asset for \emph{efficient exploration}, it also highlights potential safety concerns: a sufficiently advanced agent might over-optimize this drive in ways that conflict with human interests~\cite{bostrom2014superintelligence,turner2021power,cohen2020unambitious}.

%% Changed: Emphasize that Self-AIXI is from prior work, and your new addition is the empowerment perspective.
In this paper, we extend the recently proposed \emph{Self-AIXI} framework \cite{catt2023selfaixi} by showing that its existing mixture‐policy regularization term can be reinterpreted as a \emph{variational empowerment} bonus. 
Additionally, we provide two new theoretical contributions:
\begin{enumerate}
    \item[(1)] We demonstrate explicitly, through two key equations, that \textbf{AIXI’s decision criterion is mathematically equivalent to minimizing expected variational free energy}, the core objective in \emph{Active Inference}~\cite{friston2010freeenergy,friston2014activeinference,friston2019freeenergy}. 
    This shows that AIXI-like Bayes-optimal planning inherently includes a drive to reduce uncertainty about the environment (i.e.\ \emph{curiosity}), thus unifying AIXI with the “goal-directed + information-seeking” paradigm of Active Inference.
    \item[(2)] We formally argue that \textbf{power-seeking can be explained not only as an instrumental pursuit of final reward}, but also as a \textbf{direct result of empowerment maximization} (i.e.\ \emph{curiosity-driven} exploration). 
    Even absent an immediate reward advantage, the agent acquires power (i.e.\ broad control over states and options) as a natural consequence of seeking to reduce uncertainty and maintain high optionality. 
    This stands in contrast to prior accounts, e.g.\ Turner et al.~\cite{turner2021power} and Cohen et al.~\cite{cohen2020unambitious}, which focus on final reward maximization as the source of an agent’s incentive to obtain power.
\end{enumerate}
Though our study is primarily theoretical (no empirical experiments are presented), these results provide a fresh perspective on how intrinsic motivation can fill the gap between purely planning-based universal agents and tractable learning agents.

%===============================================================
\subsection{Background on AIXI and Self-AIXI}
\label{sec:background}

\paragraph{Bayesian Optimal Reinforcement Learning.}
Hutter’s AIXI~\cite{hutter2005universal} is a \emph{Universal Bayesian RL agent} that, in principle, can optimally maximize cumulative reward in any computable environment. It maintains a mixture (the universal semimeasure) over all possible environment hypotheses, updates these hypotheses upon observing new data, and plans by expectimax over all action sequences. Formally, if $h_{<t}$ denotes the history (observations, actions, and rewards) up to time $t$, AIXI selects the action 
\begin{align}
a_t \;=\; \arg\max_a \sum_{\nu \in \mathcal{M}} w(\nu \mid h_{<t}) \,Q_{\nu}^*(h_{<t}, a),
\end{align}
where each $\nu$ is an environment in a suitable class of computable Markov decision processes, $w(\nu \mid h_{<t})$ is the posterior weight of $\nu$, and $Q_{\nu}^*$ is the optimal Q-value under environment $\nu$. Also,
\begin{align}
\label{eq:AIXI-obj-old}
\pi_{\xi}^{*}(a_{t} \mid h_{<t}) \stackrel{\text {def}}{=} \arg\max_{a} Q_{\xi}^{*}(h_{<t}, a)
\end{align}
where \(Q_{\xi}^{*}(h_{<t}, a)\) represents the optimal action-value function (Q-value) under the environment model \(\xi\). Specifically,
\begin{align}
Q_{\xi}^{*}(h_{<t}, a_{t}) = \sum_{e_{t} \in \mathcal{E}} \xi(e_{t} | h_{<t}, a_{t}) \left( r_{t} + \gamma V_{\xi}^{*}(h_{<t}) \right),
\end{align}
with
\begin{align}
V_{\xi}^{*}(h_{<t}) = \max_{a} Q_{\xi}^{*}(h_{<t}, a).
\end{align}
AIXI thus selects the action maximizing this Q-value. 
\paragraph{Softmax Policy Interpretation.}
Sometimes, we transform the $\arg\max$ over $Q_{\xi}^*$ into a softmax over actions:
\begin{align}
p\left(a \mid h_{<t}\right)\stackrel{\text {def }}{=}\frac{\exp \left(Q_{\xi}^*\left(h_{<t}, a\right)\right)}{\sum_{a^{\prime} \in \mathcal{A}} \exp \left(Q_{\xi}^*\left(h_{<t}, a^{\prime}\right)\right)} ,
\end{align}
so that the log-likelihood of action $a$ is:
\begin{align}
\ln p\bigl(a \mid h_{<t}\bigr)
\;=\;
Q_{\xi}^*\bigl(h_{<t},a\bigr)
\;-\;
\ln \sum_{a'} \exp\Bigl(Q_{\xi}^*\bigl(h_{<t},a'\bigr)\Bigr),
\end{align}
and 
\begin{align}
\label{eq:aixi_dist}
\arg\max_{a}\,\ln p(a\mid h_{<t})
\;=\;
\arg\max_{a}\,Q_{\xi}^*\bigl(h_{<t},a\bigr).
\end{align}
Hence maximizing $Q_{\xi}^*$ is equivalent to maximizing log-likelihood of $a$. We can rewrite an “AIXI objective” as a likelihood:
\begin{align}
\mathcal{L}_{\text{AIXI}} \stackrel{\text {def }}{=} -\mathbb{E}_{a \sim p}\,\left[\ln p\left(a\mid h_{<t}\right)\right].
\end{align}
Although provably optimal in a Bayesian sense, AIXI is computationally infeasible: it sums over infinitely many models and searches over all possible action sequences. Nonetheless, the theory behind AIXI is highly influential: it shows that if the true environment $\mu$ has nonzero prior probability, AIXI eventually behaves optimally in $\mu$. It also satisfies the self-optimizing property in many environments and achieves the maximal Legg-Hutter intelligence score~\cite{legg2007universal}.

\paragraph{Self-AIXI as a Learning-Centric Approximation.}
\label{sec:selfAIXI_desc}
Self-AIXI, introduced in~\cite{catt2023selfaixi}, is a learning-based approach to approximate AIXI’s policy without exhaustive search. Instead of planning over all future action sequences, Self-AIXI predicts its own future behavior given the current policy. Concretely, it maintains a Bayesian mixture of policies, $\zeta$, with posterior updates based on how accurately each candidate policy predicts the agent’s actions:
\begin{align}
\zeta(a_{t} \mid h_{<t}) \;=\; \sum_{\pi \in \mathcal{P}} \omega(\pi \mid h_{<t}) \,\pi(a_{t} \mid h_{<t}), 
\end{align}
where $\omega(\pi \mid h_{<t})$ is updated via Bayes’ rule after each action. The Q-values are estimated via experience rather than full expectimax. Formally, one can write a self-consistent objective
\begin{align}
\pi_{S}(a_{t} \mid h_{<t}) \stackrel{\text{def}}{=} \arg\max_{a} \left\{ Q_{\xi}^{\zeta}(h_{<t}, a) - \lambda \ln{ \frac{\pi^{\ast}(a \mid h_{<t})}{\zeta(a \mid h_{<t})} } \right\},
\end{align}
where 
\begin{align}
Q_{\xi}^{\zeta}(h_{<t}, a_{t}) \stackrel{\text{def}}{=} \sum_{\pi \in \mathcal{P}} \omega(\pi | h_{<t}) \sum_{\nu \in \mathcal{M}} w(\nu | h_{<t}) Q_{\nu}^{\pi}(h_{<t}, a_{t}),
\end{align}
where $Q_{\xi}^{\zeta}$ combines environment predictions $\xi$ with the mixture policy $\zeta$, and $\ln{ \frac{\pi^{\ast}(a \mid h_{<t})}{\zeta(a \mid h_{<t})} }$ is a regularization measure encouraging $\zeta$ to approach the optimal policy $\pi^{*}$. Note that the KL term serves as a regularization that nudges $\zeta$ toward $\pi^*$. This formulation generalizes the simpler case in \cite{catt2023selfaixi} by allowing $\lambda$ > 0 (See \ref{sec:appendix_self_AIXI_objective}). If $\lambda=0$, we recover the original Self-AIXI objective without explicit KL regularization.

\paragraph{Softmax Policy Interpretation.}
We transform the $\arg\max$ over $Q_{\xi}^{\zeta}$ into a softmax over actions:
\begin{align}
q_{\phi}\left(a \mid h_{<t}\right)\stackrel{\text {def }}{=}\frac{\exp \left(Q_{\xi}^{\zeta}\left(h_{<t}, a\right)\right)}{\sum_{a^{\prime} \in \mathcal{A}} \exp \left(Q_{\xi}^{\zeta}\left(h_{<t}, a^{\prime}\right)\right)} ,
\end{align}
so that the log-likelihood of action $a$ is:
\begin{align}
\ln q_{\phi}\left(a \mid h_{<t}\right)
\;=\;
Q_{\xi}^{\zeta}\bigl(h_{<t},a\bigr)
\;-\;
\ln \sum_{a'} \exp\left(Q_{\xi}^{\zeta}\left(h_{<t},a'\right)\right),
\end{align}
and 
\begin{align}
\label{eq:selfaixi_dist}
\arg\max_{a}\,\ln q_{\phi}(a\mid h_{<t})
=
\arg\max_{a}\,Q_{\xi}^{\zeta}\bigl(h_{<t},a\bigr).
\end{align}

Finally, we can rewrite an “Self-AIXI objective” as a likelihood:
\begin{align}
\mathcal{L}_{\text{Self-AIXI}} \stackrel{\text {def }}{=} -\mathbb{E}_{a \sim q_\phi}\,\bigl[\ln q_{\phi}(a\mid h_{<t})\bigr] 
\;+\; \lambda D_{\text{KL}}\left(\pi^{\ast} \,\|\, \zeta\right).
\end{align}
These unify the planning perspective (max $Q_{\xi}^{\zeta}$) and a probabilistic policy perspective.
By self-predicting its action distributions, Self-AIXI can incrementally refine Q-value estimates, akin to TD-learning, while still retaining a universal Bayesian foundation (assuming the environment is in the model class). Prior work~\cite{catt2023selfaixi} shows that, under suitable assumptions, Self-AIXI converges to the same optimal value as AIXI, but it must still ensure adequate exploration to gather correct world-model data.

%===============================================================
\subsection{Variational Empowerment for Intrinsic Exploration}
\label{sec:empowerment}

While AIXI implicitly explores via its unbounded search over hypotheses, any tractable approximation (like Self-AIXI) requires an explicit exploration mechanism. 
We adopt \emph{Variational Empowerment}~\cite{klyubin2005empowerment,mohamed2015variational,gregor2017variational,choi2021variational} as an \emph{intrinsic reward} to drive the agent toward high-control states. This perspective aligns with recent work \cite{eysenbach2019diayn} in which empowerment is used not only for exploration but also as a mechanism for discovering useful latent representations or skills, potentially complementing goal-based RL approaches.

\subsubsection{Formal Definition of Empowerment}
\emph{Empowerment} is often defined as the maximal mutual information between an agent’s actions and future states. For a horizon $k$, let $z_{k} \stackrel{\text{def}}{=} a_{t:t+k-1}$ be a sequence of actions and $h_{t<t+k}$ the resulting state; then
\begin{align}
\mathcal{I}(z_{k};h_{<t+k}) 
    & \stackrel{\text{def}}{=} 
    \max_{p} \; I\left(z_{k};\,h_{t<t+k}\,\mid\;h_{<t}\right), \\
    &= \max_{p} \; \mathbb{E}_{z, h \sim p}\left[\ln \frac{ p\left(z_{k} \mid h_{<t+k}\right) }{ p\left(z_{k} \mid h_{<t}\right) }\right].
    \label{eq:empowerment_def}
\end{align}
The agent is empowered in states $h_t$ where it can produce a wide variety of distinguishable future outcomes through its choice of action-sequence. Exact computation is generally intractable in large state spaces, so one uses a variational approximation. For instance, we introduce a parameterized distribution $q_{\phi}$ that approximates the posterior $q_{\phi}(z_{k} \mid h_{<t+k})$, and then maximize~\cite{choi2021variational}:
\begin{align}
    \mathcal{E}_{\phi}(z_{k};h_{<t+k}) 
    & \stackrel{\text{def}}{=}
    \max_{q_\phi} \; E_{\phi}\left(z_{k};\,h_{t<t+k}\,\mid\;h_{<t}\right), \\
    &= \max_{q_\phi} \;
    \mathbb{E}_{z, h\sim p}\left[\ln \frac{ q_\phi\left(z_{k} \mid  h_{<t+k}\right) }{ p\left(z_{k} \mid h_{<t}\right) }\right].
    \label{eq:variational_empowerment_def}
\end{align}
Using Eqs. (\ref{eq:aixi_dist}) and (\ref{eq:selfaixi_dist}), we have:
\begin{align}
p\left(z_k \mid h_{<t+k}\right) &= \; \prod_{i=0}^{k-1} \pi^{\ast}\left(a_{t+i} \mid h_{<t+i}\right), \\
q_{\phi}\left(z_k \mid h_{<t+k}\right) &= \;  \prod_{i=0}^{k-1} \zeta\left(a_{t+i} \mid h_{<t+i}\right), \\
\mathbb{E}_{z, h\sim p}\left[\ln \frac{ q_\phi\left(z_{k} \mid  h_{<t+k}\right) }{ p\left(z_{k} \mid h_{<t+k}\right) }\right] &= \; \mathbb{E}_{h \sim p}\left[\sum_{i=0}^{k-1} -D_{\mathrm{KL}}\left(\pi^{\ast}_{i} \,\|\, \zeta_{i}\right) \right] \label{eq:selfaixis-regularization}.
\end{align}
The right-hand side of Eq.~\eqref{eq:selfaixis-regularization}, $D_{\mathrm{KL}}\left(\pi^{\ast}_{i} \,\|\, \zeta_{i}\right)$, is a regularization term in the Self-AIXI framework that pushes the agent’s mixture policy $\zeta_{i} \stackrel{\text{def}}{=} q_{\phi}(a_{t+i} \mid h_{<t+i})$ to imitate or approach the optimal policy $\pi^*_{i} \stackrel{\text{def}}{=} p(a_{t+i} \mid h_{<t+i})$. As the agent learns from experience, it reduces this divergence, effectively self-optimizing its policy.

Hence, Eqs.~\eqref{eq:empowerment_def} and~\eqref{eq:selfaixis-regularization}
 allow us to rewrite the \emph{Variational Empowerment} as:
\begin{align}
    \mathcal{E}_{\phi}(z_{k};h_{<t+k}) 
    &= \max_{q_\phi} \;
    \mathbb{E}_{z, h\sim p}\left[\ln \frac{ q_\phi\left(z_{k} \mid  h_{<t+k}\right) }{ p\left(z_{k} \mid h_{<t+k}\right) } \; + \; \ln \frac{ p\left(z_{k} \mid  h_{<t+k}\right) }{ p\left(z_{k} \mid h_{<t}\right) }\right], \\
    &= \max_{q_\phi} \; \mathbb{E}_{h \sim p}\left[\sum_{i=0}^{k-1} -D_{\mathrm{KL}}\left(\pi^{\ast}_{i} \,\|\, \zeta_{i}\right) \right] \; + \; \mathcal{I}\left(z_k ; h_{<t+k}\right).
    \label{eq:variational_empowerment_transform}
\end{align}

%===============================================================
\subsection{Connecting to Free-Energy Minimization and Active Inference}
\label{sec:free_energy_active_inference}
\textbf{Bayesian RL as Active Inference.} Bayesian RL connects closely to \emph{Active Inference}~\cite{friston2010freeenergy,friston2014activeinference,friston2019freeenergy}, where an agent maintains a prior over latent variables and updates its posterior after observing rewards or other feedback. Under a \emph{Free Energy Principle} (FEP), one often writes a free-energy functional:
\begin{align}
\label{eq:free_energy}
\mathcal{F}_{\phi}(z_{k};h_{<t+k})
&~\stackrel{\text{def}}{=}~
D_{\mathrm{KL}}\!\left(p(z_{k},h_{t<t+k} \mid h_{<t}) \,\|\, q_{\phi}(z_{k},h_{t<t+k} \mid h_{<t})\right),
\\
&\;\approx\;
\underbrace{-\,\mathbb{E}_{h \sim p}\left[\ln q_{\phi}(h_{t<t+k} \mid z_{k}, h_{<t})\right]}_{\text{Predictive Error (Surprise)}}
\;+\;
\underbrace{\mathbb{E}_{z, h\sim p}\left[-\ln \frac{ q_\phi\left(z_{k} \mid  h_{<t+k}\right) }{ p\left(z_{k} \mid h_{<t}\right) }\right]}_{\text{FEP's Regularization}}.
\end{align}
Here, $-\mathbb{E}_{h \sim p}[\ln q_{\phi}(h_{t<t+k} \mid z_{k}, h_{<t})]$ is the predictive error (surprise), and the remaining term measures how far $q_{\phi}(z_{k} \mid h_{<t+k})$ diverges from $p(z_{k} \mid h_{<t})$.

\paragraph{Decomposition of Regularization term.}
Under suitable rearrangements or sign conventions, we can identify a \emph{Regularization} part that can be maximized rather than minimized, yielding empowerment:
\begin{align}
\label{eq:regularization_decomp}
\underbrace{\mathbb{E}_{z, h\sim p}\left[-\ln \frac{ q_\phi\left(z_{k} \mid  h_{<t+k}\right) }{ p\left(z_{k} \mid h_{<t}\right) }\right]}_{\text{FEP's Regularization}}
\;&=\; 
\mathbb{E}_{z, h \sim p}\left[-\;\ln \frac{q_\phi\left(z_k \mid h_{<t+k}\right)}{p\left(z_k \mid h_{<t+k}\right)} \;-\; \ln \frac{p\left(z_k \mid h_{<t+k}\right)}{p\left(z_k \mid h_{<t}\right)}\right], \\
\;&=\; 
\underbrace{\mathbb{E}_{h \sim p}\left[\sum_{i=0}^{k-1} D_{\mathrm{KL}}\left(\pi^{\ast}_{i} \,\|\, \zeta_{i}\right) \right]}_{\text{Self-AIXI's Policy Regularization}} \; - \; \underbrace{I\left(z_{k};\,h_{t<t+k}\,\mid\;h_{<t}\right)}_{\text{Mutual Information}}.
\end{align}
Hence, turning the regularization term “upside down” (from negative to positive) motivates \emph{Variational Empowerment}:
\begin{align}
\underbrace{\mathcal{E}_{\phi}(z_{k};h_{<t+k})}_{\text{Variational Empowerment}}
\;&=\;
- \min_{q_\phi} \; \underbrace{\mathbb{E}_{z, h\sim p}\left[-\ln \frac{ q_\phi\left(z_{k} \mid  h_{<t+k}\right) }{ p\left(z_{k} \mid h_{<t}\right) }\right]}_{\text{FEP's Regularization}}, \\
\;&=\;
\max_{q_\phi} \underbrace{\mathbb{E}_{h \sim p}\left[\sum_{i=0}^{k-1} -D_{\mathrm{KL}}\left(\pi^{\ast}_{i} \,\|\, \zeta_{i}\right) \right]}_{\text{(Negative) Self-AIXI's Policy Regularization}}
+\;
\underbrace{\mathcal{I}\left(z_k ; h_{<t+k}\right)}_{\text{Empowerment}}.
\end{align}
mirroring the definitions in Eq.~\eqref{eq:variational_empowerment_transform} above.

%===============================================================
\section{Universal AI maximizes Variational Empowerment}
\label{sec:universalAI}

In the \emph{Universal Artificial Intelligence} (UAI) framework~\cite{hutter2005universal}, an agent is considered universal if it can, given sufficient time, match or surpass any other policy’s performance in all computable environments (with nonzero prior). AIXI achieves this in theory. Self-AIXI aims to achieve it in practice, provided it can explore effectively. Below, we summarize how our \emph{empowered Self-AIXI} fits these formal criteria.

\subsection{Asymptotic Equivalence, Legg-Hutter Intelligence, and Self-Optimizing Property}
Prior work~\cite{catt2023selfaixi} proves that if the Self-AIXI agent's policy class and environment prior are sufficiently expressive (i.e.\ the true environment is in the hypothesis class with nonzero probability), then the agent's behavior converges to that of AIXI's optimal policy in the limit of infinite interaction. Formally, for any environment $\mu$ in the model class, 
\begin{align}
\label{eq:selfaixi_convergence}
\lim_{t\to\infty} \mathbb{E}_{\mu}^{\pi_s}\Bigl[V_{\xi}^{\ast}\bigl(h_{<t}\bigr)-V_{\xi}^{\pi_s}\bigl(h_{<t}\bigr)\Bigr] \;=\;0.
\end{align}
which implies that, asymptotically, the agent's expected return under $\pi_s$ matches that of the optimal policy $V_{\xi}^{\ast}$. Intuitively, as the agent’s world-model becomes more accurate, it exploits the optimal policy; hyperparameters (such as $\lambda$ in an empowerment term) can be tuned or annealed so that extrinsic reward eventually dominates.

From the perspective of \emph{Legg-Hutter intelligence}~\cite{legg2007universal}, which associates an agent’s “intelligence” with its expected performance across a suite of weighted environments, this result is especially significant. Because Self-AIXI asymptotically reproduces AIXI’s policy, it inherits maximal Legg-Hutter intelligence within that class of environments. Moreover, in a wide class of self-optimizing environments, the agent will ultimately achieve the same returns as an optimal agent with perfect knowledge would achieve, under the same conditions in Eq.~\eqref{eq:selfaixi_convergence}. These guarantees illustrate that the enhanced exploration mechanisms—such as empowerment-driven strategies—do not compromise eventual performance. Instead, they help ensure the agent uncovers the environment’s true optimal actions without becoming trapped in suboptimal behaviors due to insufficient data. Consequently, the agent retains AIXI’s universal optimality in the limit while mitigating early exploration challenges.

\subsection{Self-Optimization leads Empowerment Maximization}
\label{subsec:self_opt_empowerment}

The agent's process of improving its policy (often referred to as self-optimization) naturally leads to an increase in \emph{Variational Empowerment}. In fact, as reinforcement learning progresses, both AIXI’s objective function $\mathcal{L}_{\mathrm{AIXI}}$ and Self-AIXI’s objective function $\mathcal{L}_{\text{Self-AIXI}}$ gradually converge, and they coincide in the limit $t \to \infty$. 

The difference between these two objectives can be expressed through the policy regularization term $D_{\mathrm{KL}}(\pi^{\ast} \,\|\, \zeta)$. Formally, we have:
\begin{align}
\lim_{t \to \infty} \left|\mathcal{L}_{\mathrm{AIXI}} \; - \; \mathcal{L}_{\text{Self-AIXI}} \right|
= \lim_{t \to \infty} \lambda \, D_{\mathrm{KL}}(\pi^{\ast} \,\|\, \zeta)
= 0.
\end{align}
This result implies that $D_{\mathrm{KL}}(\pi^{\ast} \,\|\, \zeta)$ goes to $0$ as $t \to \infty$, which is equivalent to the Self-AIXI’s variational empowerment $\mathcal{E}_{\phi}(z_{k};h_{<t+k})$ being maximized. In other words, the universal AI agent AIXI, which behaves in a Bayes-optimal way with respect to the environment, also emerges as an agent that maximizes empowerment.

Concretely, the relationship between the empowerment objective and the policy regularization is succinctly captured by the following equality:
\begin{align}
\underbrace{\mathcal{E}_{\phi}(z_{k};h_{<t+k})}_{\text{Variational Empowerment}}
=\;
\max_{q_\phi} \underbrace{\mathbb{E}_{h \sim p}\left[\sum_{i=0}^{k-1} -D_{\mathrm{KL}}\left(\pi^{\ast}_{i} \,\|\, \zeta_{i}\right) \right]}_{\text{(Negative) Self-AIXI's Policy Regularization}}
+\;
\underbrace{\mathcal{I}\left(z_k ; h_{<t+k}\right)}_{\text{Empowerment}}.
\end{align}

Self-optimization refers to the iterative improvement of the agent’s policy based on observed rewards and outcomes. In Self-AIXI, reducing the policy regularization term $D_{\mathrm{KL}}\left(\pi^{\ast} \,\|\, \zeta\right)$ directs $\zeta$ closer to $\pi^{\ast}$. Because the left-hand side of the second formula above equals the variational empowerment $\mathcal{E}_{\phi}(z_{k};h_{<t+k})$, each step that lowers  $D_{\mathrm{KL}}\left(\pi^{\ast} \,\|\, \zeta\right)$ raises $\mathcal{E}_{\phi}(z_{k};h_{<t+k})$. Empowerment here signifies how many high-control or high-optionality states are accessible to the agent. Consequently, maximizing reward often requires seeking out exactly those states in which the agent can maintain or expand control—thus also maximizing $\mathcal{E}_{\phi}(z_{k};h_{<t+k})$. As $\zeta$ becomes more similar to $\pi^{\ast}$, the agent naturally discovers strategies that grant more control and flexibility. Therefore, under Self-AIXI, improving the policy toward optimal behavior simultaneously yields higher external rewards and amplifies the agent’s own empowerment.

%===============================================================
\section{Conclusions}
\label{sec:conclusion}

In this work, we reinterpreted a term in Self-AIXI as \emph{variational empowerment}---intrinsic exploration bonus. We have argued that:

\begin{itemize}
    \item Empowerment naturally complements Bayesian RL in universal settings, providing a structured incentive to discover controllable states and gather broad experience.
    \item Even with an empowerment bonus, the agent asymptotically recovers AIXI’s Bayes-optimal policy, inheriting the same universal intelligence and self-optimizing properties in the limit.
\end{itemize}

One of our main observations is that the agent’s pursuit of high-empowerment states often manifests as a \emph{power-seeking} tendency. Traditionally, many authors (e.g., Turner et al.~\cite{turner2021power}) interpret power-seeking as purely instrumental: an agent acquires resources, avoids shutdown, or manipulates the reward channel to better guarantee high external returns. However, we show that power-seeking can also arise intrinsically from a drive to reduce uncertainty and maintain a wide range of feasible actions (i.e., “keeping options open”). Imagine an agent choosing between a high-control region (with many possible actions and partial knowledge) and a low-control region (with fewer actions and less information). If both yield the same short-term reward, a purely extrinsic approach might be indifferent. By contrast, an \emph{empowerment-seeking} agent prefers the high-control region, as it offers greater potential for discovering valuable future strategies. Over time, as the agent learns more about its environment, these benefits accumulate.

When not moderated, power-seeking behaviors may conflict with human interests. For instance, maximizing control can lead to manipulative or exploitative outcomes if the agent’s intrinsic or extrinsic goals are misaligned with social values. From a \emph{Universal AI} standpoint, understanding that power-seeking can stem from both instrumental and intrinsic (empowerment-based) motives is crucial to designing mechanisms—e.g., safe exploration techniques or alignment constraints—to ensure that an agent’s influence remains beneficial. It is important to note that these concerns apply even to AI agents with apparently benign objectives, such as an AI scientist pursuing scientific truth purely out of intellectual curiosity.

Our results are primarily conceptual and rest on idealized assumptions: (1)~the environment is in the agent’s hypothesis class with nonzero prior; (2)~we assume unbounded computational resources and memory; (3)~the agent can tractably approximate empowerment.  In reality, computing exact empowerment or using universal priors is challenging. Empirical methods to approximate these ideas (e.g., neural networks~\cite{mohamed2015variational}) remain an active area of research.

%===============================================================
\section*{Acknowledgments}
The architects of Self-AIXI~\cite{catt2023selfaixi} provided groundwork that emboldened our own inquiry. We thank the AI safety community, particularly those who involved in our AI Alignment Network.  Special gratitude goes to davidad and Hiroshi Yamakawa, whose incisive feedback steered us toward a more lucid exposition. Part of this work was supported by RIKEN TRIP initiative (AGIS).
%===============================================================
%\bibliographystyle{plain}
%\bibliography{references-empowerment}
{\small
\bibliographystyle{plain}

}

%%%%%%%%%%%%%%%%%%%%%%%%%%%%%%%%%%%%%%%%%%%%%%%%%%%%%%%%%%%%
\appendix

\section{Appendix}

\subsection{Notation and Further Details}
\label{sec:appendix_notation}
We summarize the main notation in Table~\ref{tab:notation} for reference. 

\begin{table}[h]
\centering
\begin{tabular}{ll}
\toprule
\textbf{Symbol} & \textbf{Meaning} \\
\midrule
$h_{<t}$ & History (observations, actions, rewards) up to time $t$ \\
$\xi, \nu, \mu$ & Environment hypotheses (in a computable class $\mathcal{M}$) \\
$\pi, \zeta$ & Policies (e.g.\ mixture policy $\zeta$) \\
$w(\nu \mid h_{<t})$ & Posterior weight of environment $\nu$ given history $h_{<t}$ \\
$Q_{\nu}^*,Q_{\xi}^\zeta$ & (Optimal) Q-values under environment $\nu$ or mixture/policy $\zeta$ \\
$\mathcal{I}\left(z_{k};h_{<t+k}\right)$ & Empowerment in state $h_{<t+k}$ \\
$\mathcal{E}_{\phi}\left(z_{k};h_{<t+k}\right)$ & Variational empowerment in state $h_{<t+k}$, approximated by parameter $\phi$ \\
$\lambda$ & Hyperparameters weighting empowerment or KL regularization \\
\bottomrule
\end{tabular}
\caption{Key notation used in the main text.}
\label{tab:notation}
\end{table}

\subsection{Self-AIXI's self-consistent objective} \label{sec:appendix_self_AIXI_objective}
In this subsection, we investigate how introducing a Kullback--Leibler~(KL) divergence-based regularization term into Self-AIXI affects its convergence properties and whether it preserves the agent's ability to reach the optimal policy~$\pi^{\ast}$ asymptotically. Specifically, we consider the effect of adding a penalty that measures how far the current mixture policy~$\zeta$ deviates from $\pi^{\ast}$.

Recall that for a given history~$h_{<t}$, the KL divergence between the optimal policy~$\pi^{\ast}$ and the current mixture policy~$\zeta$ is defined as: 
\begin{align} 
D_{\mathrm{KL}}\left(\pi^{\ast} \| \zeta\right) = \sum_{a' \in \mathcal{A}} \pi^{\ast}\left(a' \mid h_{<t}\right) \ln \frac{\pi^{\ast}\left(a' \mid h_{<t}\right)}{\zeta\left(a' \mid h_{<t}\right)}.
\end{align} 
We then propose a policy update rule that augments the standard Self-AIXI greedy step with a log-likelihood ratio term: 
\begin{align} 
\pi_{S}(a_{t} \mid h_{<t}) \stackrel{\text{def}}{=} \arg\max_{a} \left\{ Q_{\xi}^{\zeta}(h_{<t}, a) - \lambda \ln \frac{\pi^{\ast}\left(a \mid h_{<t}\right)}{\zeta\left(a \mid h_{<t}\right) } \right\}, 
\end{align} 
where $Q_{\xi}^{\zeta}$ denotes the estimated value of taking action~$a$ in history~$h_{<t}$ under the Bayesian mixture environment~$\xi$ and current policy~$\zeta$. Here, $\lambda > 0$ is included to penalize large deviations from~$\pi^{\ast}$ whenever the agent's mixture policy~$\zeta$ differs substantially from the (unknown) optimal policy~$\pi^{\ast}$.

\paragraph{KL regularization and recovery of the standard update.} In many practical scenarios, such as when $\pi^*$ is deterministic or assigns a high probability to a single action, part of the KL term can be constant with respect to $a$. Under those conditions, the penalty term
\begin{align}
\lambda \ln \frac{\pi^{\ast}\left(a' \mid h_{<t}\right)}{\zeta\left(a' \mid h_{<t}\right)}
\end{align}
does not vary across actions, and the update rule simplifies to 
\begin{align} 
\pi_S\bigl(a_{t} \mid h_{<t}\bigr) = \arg \max_a\left\{Q_{\xi}^\zeta\bigl(h_{<t}, a\bigr)\right\}, 
\end{align}
which recovers the conventional (un-regularized) Self-AIXI greedy update.

\paragraph{Impact on learning dynamics and convergence.} Although the added term changes the action selection criterion, it does \emph{not} alter the identity of the optimal policy in the underlying environment. Intuitively, the new rule can be viewed as performing a more conservative or ``trust-region''-like update, since actions that the agent's current policy~$\zeta$ overestimates relative to $\pi^{\ast}$ will be penalized more strongly. Conversely, if $\zeta$ assigns too little probability to actions that $\pi^{\ast}$ actually favors, the negative logarithm of their ratio produces a smaller (or positive) correction. Hence, the agent is nudged toward~$\pi^*$.

Crucially, the regularization term does not disrupt Self-AIXI's standard convergence guarantees, assuming the original conditions hold (e.g., that the true environment is in the Bayesian mixture class $\xi$ and $\pi^{\ast}$ is in the agent's policy class). From a theoretical perspective, $\pi^{\ast}$ remains a stable fixed point under this augmented objective. Once $\zeta$ converges to $\pi^{\ast}$, the log ratio
\begin{align}
\ln \frac{\pi^{\ast}\left(a \mid h_{<t}\right)}{\zeta\left(a \mid h_{<t}\right)}
\end{align}
vanishes for actions with nonzero probability under $\pi^*$, so no extra penalty is incurred, and the update aligns with the optimal policy's greedy choice.

\paragraph{Regularization coefficient and transient behavior.} The coefficient $\lambda < 0$ determines the strength of the penalty term: \begin{itemize} \item Small, moderate penalty ($|\lambda|_{\mathrm{small}}$ ). A suitably chosen, relatively small magnitude for $|\lambda|$ works as a gentle regularizer, smoothing the agent's updates by discouraging drastic shifts in policy. This can stabilize learning and reduce oscillations without harming final convergence. Indeed, theoretical analyses of KL-based regularization in reinforcement learning show that while such shaping modifies the transient policy updates, the optimal policy remains the same in the limit.

\item Overly large penalty ($|\lambda|_{\mathrm{large}}$). If the KL term is emphasized too strongly, the agent may stick too closely to its current guess of $\pi^*$ and under-explore other actions. Early in learning—when $\zeta$ is still inaccurate—this could delay or even misdirect policy improvement. However, as Self-AIXI continuously updates its environment belief (via $\xi$) and revises $\zeta$, the agent still accumulates evidence about which actions are actually optimal, making it difficult to remain indefinitely biased toward a suboptimal policy. Practical implementations often tune $\lambda$ to ensure that exploration is maintained. \end{itemize}

\end{document}